\documentclass[10pt, twocolumn]{article}

\usepackage[T1]{fontenc}
\usepackage[utf8]{inputenc}
\usepackage{lmodern}
\usepackage{times}
\usepackage{soul}

\usepackage[
  top=25mm,
  bottom=25mm,
  left=15mm,
  right=15mm,
  columnsep=6mm
]{geometry}

\usepackage[
  activate={true,nocompatibility},
  final,
  tracking=true,
  kerning=true,
  spacing=true,
  factor=1100,
  stretch=10,
  shrink=10
]{microtype}
\microtypecontext{spacing=nonfrench}

\usepackage{amsmath}
\usepackage{amssymb}
\usepackage{amsfonts}
\usepackage{mathtools}

\usepackage{graphicx}
\usepackage{xcolor}

\usepackage{booktabs}
\usepackage{multirow}
\usepackage{tabularx}

\usepackage{placeins}
\usepackage{stfloats}
\usepackage{caption}
\usepackage{subcaption}

\usepackage{enumitem}

\usepackage{algorithm}
\usepackage{algpseudocode}

\usepackage{tikz}
\usetikzlibrary{
  shapes.geometric,
  arrows.meta,
  positioning,
  fit,
  backgrounds,
  calc
}

\usepackage{titling}
\usepackage{abstract}

\usepackage{titlesec}

\usepackage[numbers, sort&compress]{natbib}

\usepackage[colorlinks=true]{hyperref}
\hypersetup{
  linkcolor = blue!70!black,
  citecolor = blue!70!black,
  urlcolor  = blue!70!black,
}

\titleformat{\section}
  {\large\bfseries}{\thesection.}{0.5em}{}
\titleformat{\subsection}
  {\normalsize\bfseries}{\thesubsection.}{0.5em}{}
\titleformat{\subsubsection}
  {\normalsize\itshape}{\thesubsubsection.}{0.5em}{}

\pretitle{%
  \begin{center}
  \rule{\linewidth}{1.5pt}\\[8pt]
  \Large\bfseries\scshape
}
\posttitle{%
  \end{center}
  \vskip5pt
  \noindent\rule{\linewidth}{0.4pt}\\[-2pt]
  \noindent\rule{\linewidth}{1.5pt}
  \vskip10pt
  \begin{center}{\small\scshape A Preprint}\end{center}
  \vskip14pt
}
\preauthor{\begin{center}\normalsize}
\postauthor{%
  \end{center}
  \vskip4pt
  \begin{center}
    \footnotesize $^{*}$Corresponding author.
  \end{center}
}
\predate{}\postdate{}

\title{%
  Gated-SwinRMT: Unifying Swin Windowed Attention with\\
  Retentive Manhattan Decay via Input-Dependent Gating%
}

\author{%
  \begin{minipage}[t]{0.30\linewidth}
    \centering
    \textbf{Dipan Maity}$^{*}$\\[3pt]
    Student\\
    Kolkata, West Bengal, India\\[2pt]
    \texttt{dipanai.xyz@gmail.com}
  \end{minipage}
  \hfill
  \begin{minipage}[t]{0.34\linewidth}
    \centering
    \textbf{Suman Mondal}\\[3pt]
    Department of Computer Science\\
    Yogoda Satsanga Palpara Mahavidyalaya\\
    West Bengal, India\\[2pt]
    \texttt{asuman.mondal2014@gmail.com}
  \end{minipage}
  \hfill
  \begin{minipage}[t]{0.30\linewidth}
    \centering
    \textbf{Arindam Roy}\\[3pt]
    Department of Computer Science \& Application\\
    Prabhat Kumar College Contai\\
    West Bengal, India\\[2pt]
    \texttt{arindamr@pkcollegecontai.ac.in}
  \end{minipage}%
}

\date{}

\begin{document}

\twocolumn[%
  \maketitle
  \vskip6mm
]


\begin{abstract}
We introduce \textbf{Gated-SwinRMT}, a family of hybrid vision
transformers that combines the shifted-window attention of the Swin
Transformer~\cite{liu2021swin} with the Manhattan-distance spatial decay
of Retentive Networks (RMT)~\cite{fan2023rmt}, augmented by
input-dependent gating. Self-attention is decomposed into consecutive
width-wise and height-wise retention passes within each shifted window,
where per-head exponential decay masks provide a two-dimensional locality
prior without learned positional biases.
Two variants are proposed.
\textbf{Gated-SwinRMT-SWAT} substitutes softmax with sigmoid activation,
implements balanced ALiBi slopes with multiplicative post-activation
spatial decay, and gates the value projection via SwiGLU; the
normalized output implicitly suppresses uninformative attention scores.
\textbf{Gated-SwinRMT-Retention} retains softmax-normalized retention
with an additive log-space decay bias and incorporates an explicit G1
sigmoid gate---projected from the block input and applied after local
context enhancement (LCE) but prior to the output
projection~$W_O$---to alleviate the low-rank $W_V \!\cdot\! W_O$
bottleneck and enable input-dependent suppression of attended outputs.
// We assess both variants on Mini-ImageNet ($224{\times}224$, 100 classes)
and CIFAR-10 ($32{\times}32$, 10 classes) under identical training
protocols, utilizing a single GPU due to resource limitations. At
${\approx}77$--$79$\,M parameters, Gated-SwinRMT-SWAT achieves
$80.22\%$ and Gated-SwinRMT-Retention $78.20\%$ top-1 test accuracy on
Mini-ImageNet, compared with $73.74\%$ for the RMT baseline. On
CIFAR-10---where small feature maps cause the adaptive windowing
mechanism to collapse attention to global scope---the accuracy advantage
compresses from $+6.48$\,pp to $+0.56$\,pp. 
\end{abstract}

\vspace{1em}\hrule\vspace{1em}
\section{Introduction}
\label{sec:introduction}

\begin{figure*}[!ht]
  \centering
  \includegraphics[width=1.0\textwidth, height=0.95\textheight, keepaspectratio]{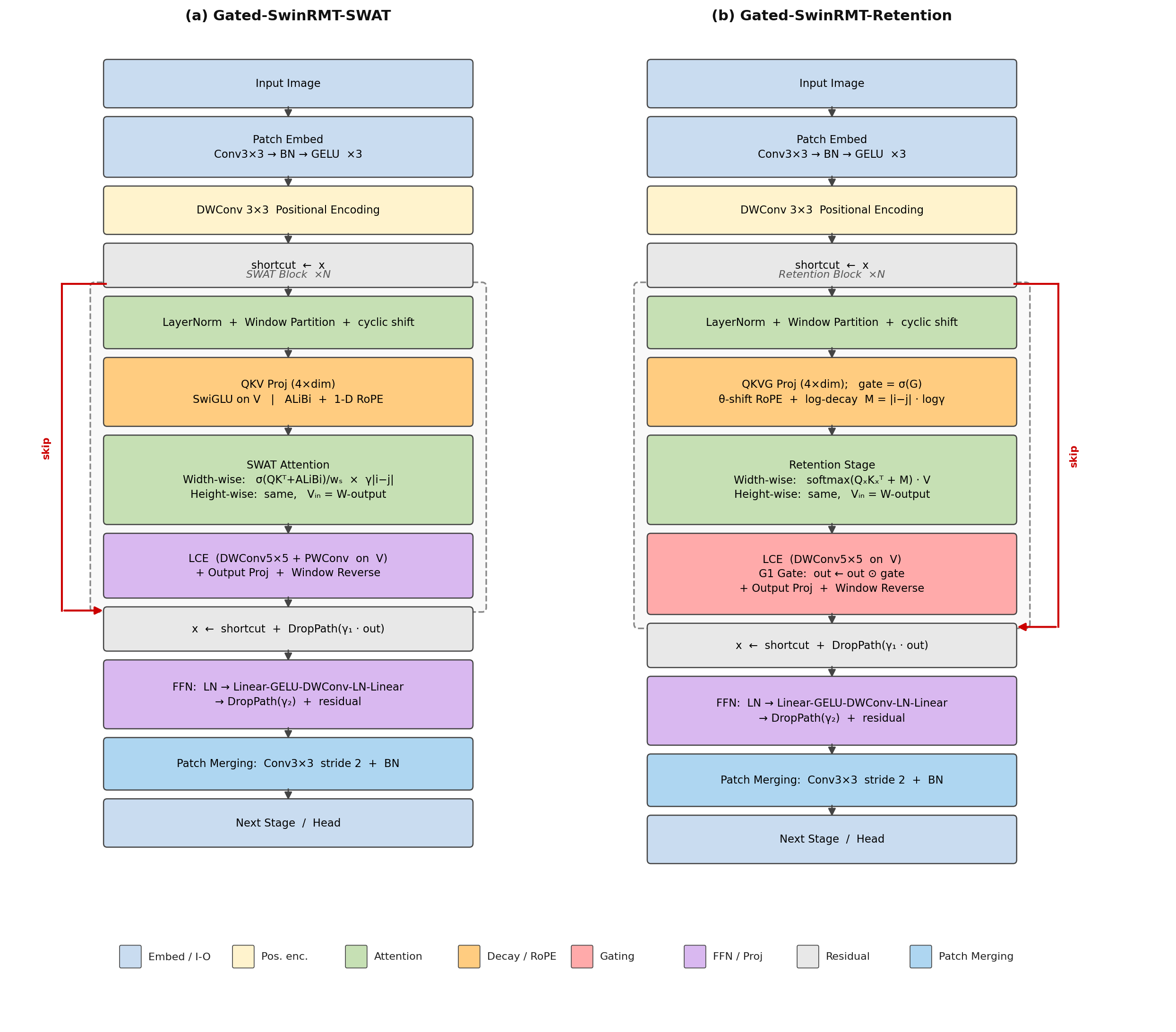}
  \caption{%
    Architecture of the proposed \textbf{Gated-SwinRMT} variants.
    \textbf{(a) Gated-SwinRMT-SWAT}: sigmoid-based normalized
    attention with SwiGLU-gated values, balanced ALiBi positional
    bias, and multiplicative spatial decay $\gamma^{|i-j|}$ applied
    post-sigmoid.
    \textbf{(b) Gated-SwinRMT-Retention}: softmax-normalized
    retention with additive log-decay mask
    $M_{ij}{=}|i{-}j|\log\gamma_h$ applied pre-softmax, and a
    learned G1 sigmoid gate applied after local context enhancement
    (LCE) and before the output projection~$W_O$.
    Both variants share DWConv\,$3{\times}3$ positional encoding,
    LayerScale ($\gamma_1, \gamma_2$) with DropPath, an LCE module
    on~$V$, adaptive window partitioning with optional cyclic shift,
    and convolutional patch merging.
    Best viewed in colour.%
  }
  \label{fig:arch}
\end{figure*}

Vision Transformers~(ViTs) have become competitive backbones for image
recognition, yet their core self-attention mechanism carries two
well-known limitations: quadratic cost in the number of spatial tokens,
and the absence of an explicit spatial prior---all token pairs receive
equal treatment regardless of distance, leaving locality entirely to
position encodings and data.

\textbf{Swin Transformer}~\cite{liu2021swin} addressed the efficiency
problem by confining attention to fixed-size non-overlapping windows and
alternating between regular and shifted partitions to propagate
information across boundaries. The resulting linear-complexity
hierarchical pyramid brought Transformers to parity with convolutional
networks on dense prediction tasks. However, Swin encodes spatial
locality only through window boundaries and a learned relative position
bias; no principled distance-weighted decay modulates the attention
weights themselves.

\textbf{RMT}~\cite{fan2023rmt} addressed the spatial-prior gap by
extending the exponential decay of RetNet~\cite{sun2023retnet} to 2-D
images. Manhattan Self-Attention~(MaSA) multiplies each attention score
by $\gamma^{|d|}$, where $|d|$ is the Manhattan distance between tokens,
encoding locality by construction. To preserve linear complexity, RMT
decomposes 2-D attention into sequential width-wise and height-wise 1-D
retention passes governed by log-space decay masks.

\paragraph{The windowed-softmax problem.}
Fusing these two designs is natural but exposes a third difficulty.
Softmax forces attention weights to sum to unity within every window,
compelling the model to attend to \emph{something} in each local
neighborhood regardless of whether any token there is informative. Under
RMT's global attention this is benign---weight redistributes across the
full feature map---but within a small window the model cannot compensate.
We observe a symptom consistent with this analysis: our ungated softmax
Retention variant loses ${\approx}6$\,pp when moving from CIFAR-10 (where
small feature maps cause the window to span the entire spatial extent,
making windowing trivial) to Mini-ImageNet at $224{\times}224$ (where
early stages operate with genuine sub-feature-map windows). The same
deficit is absent in the sigmoid-based SWAT variant, whose normalized
scores are not subject to this constraint.

\paragraph{Gated attention as a remedy.}
Recent work on gated attention in large language
models~\cite{qiu2025gatedattentionlargelanguage} shows that a learned
sigmoid gate placed after value aggregation and before the output
projection~$W_O$ can break the low-rank $W_V \!\cdot\! W_O$ bottleneck
and introduce input-dependent sparsity, allowing the model to suppress
an entire head's output when the retrieved content is uninformative. We
hypothesis that an analogous mechanism can mitigate the
windowed-softmax pathology described above.

\paragraph{Contributions.}
We introduce \textbf{Gated-SwinRMT}, a family of hybrid vision
transformers that combines Swin's hierarchical shifted-window structure
with RMT's Manhattan-distance spatial decay and input-dependent gating.
Self-attention is decomposed into consecutive width-wise and height-wise
retention passes within shifted windows, with per-head exponential decay
masks encoding 2-D locality without learned positional biases. An
adaptive windowing strategy clamps the effective window to the feature-map
size at low resolutions, allowing windowed attention to degrade gracefully
to global attention. We propose two variants:

\begin{itemize}[leftmargin=*, itemsep=3pt]
  \item \textbf{Gated-SwinRMT-SWAT} uses sigmoid activation with
    balanced ALiBi slopes, multiplicative post-activation spatial decay,
    and a SwiGLU value gate. The normalized sigmoid output provides
    implicit suppression of uninformative scores, making an explicit
    output gate unnecessary.

  \item \textbf{Gated-SwinRMT-Retention} uses softmax-normalized
    decomposed retention with additive log-space pre-normalization
    decay, and adds an explicit G1 sigmoid gate---projected from the
    block input and applied after local context enhancement~(LCE) but
    before~$W_O$---to recover selective suppression that softmax cannot
    provide.
\end{itemize}

\noindent We benchmark both variants against a pure RMT baseline at
matched parameter budgets (${\approx}77$--$79$\,M) on Mini-ImageNet and
CIFAR-10 under identical training conditions using a single GPU. The
results are consistent with the hypothesis that gated attention
mitigates the windowed-softmax pathology: the proposed variants
outperform the RMT baseline by up to $+6.48$\,pp on Mini-ImageNet,
while the advantage compresses to $+0.56$\,pp on CIFAR-10 where
windowing is effectively bypassed. We note that these conclusions rest
on indirect ablations comparing complete models, and that validation at
full ImageNet-1k scale remains future work.
\section{Method}

\subsection{Preliminary}

\paragraph{Swin Transformer.}
Given an input feature map $\mathbf{X} \in \mathbb{R}^{B \times H \times W \times C}$,
the Swin Transformer~\cite{liu2021swin} partitions the spatial domain into
non-overlapping windows of fixed size $M \times M$, yielding
$\lceil H/M \rceil \times \lceil W/M \rceil$ windows each containing
$M^2$ tokens.
Self-attention is computed independently within each window, reducing the
complexity of global self-attention from $\mathcal{O}(H^2 W^2)$ to
$\mathcal{O}(M^2 HW)$.
To enable cross-window information exchange, consecutive layers alternate
between a \emph{regular} partition and a \emph{shifted} partition, where
the grid is displaced by $(\lfloor M/2 \rfloor, \lfloor M/2 \rfloor)$
pixels before windowing and a cyclic-shift masking strategy restores
efficient batch computation.
A four-stage hierarchical design progressively halves the spatial resolution
while doubling the channel dimension, producing multi-scale feature
representations suitable for downstream dense prediction.

\paragraph{Decomposed Manhattan Self-Attention (DMSA).}
RMT~\cite{fan2023rmt} adapts the retention mechanism of
RetNet~\cite{sun2023retnet} to vision by factorising two-dimensional
spatial attention into two sequential one-dimensional passes.
For a window of tokens indexed $(i, j)$, the \emph{width-wise} pass
computes attention along the horizontal axis for each row independently,
and its output serves as the value input for the \emph{height-wise} pass
along the vertical axis.
Formally, the retention score between positions $i$ and $j$ along a single
axis is weighted by an exponential spatial decay:
\begin{equation}
  S_{ij} = \gamma^{|d|}, \qquad d = i - j,
  \label{eq:decay}
\end{equation}
where $\gamma \in (0,1)$ is a per-head learnable decay rate and
$|d|$ is the Manhattan distance along the current axis.
This factorized decomposition preserves the $\mathcal{O}(M^2)$ window
complexity of Swin while introducing an implicit inductive bias toward
local spatial coherence through the decay in \eqref{eq:decay}.

\section{SwinRMT}

\subsection{Decay Placement: Before vs.\ After Softmax}
\label{sec:decay_bug}

A subtle but consequential implementation choice concerns \emph{where} the
exponential decay $\gamma^{|d|}$ is applied relative to the softmax
normalization.

\paragraph{Multiplicative post-softmax decay (incorrect).}
The naive formulation multiplies the decay directly onto the softmax output:
\begin{equation}
  \mathbf{A}^{\text{mult}}_{ij}
  = \frac{\exp(q_i k_j^\top / \sqrt{d})}
         {\sum_l \exp(q_i k_l^\top / \sqrt{d})}
  \cdot \gamma^{|i-j|}.
  \label{eq:mult_decay}
\end{equation}
Equation~\eqref{eq:mult_decay} violates row-stochasticity: the rows of
$\mathbf{A}^{\text{mult}}$ no longer sum to one, which causes the
effective attention mass to shrink with distance and leads to gradient
instability for long sequences.
Moreover, the multiplicative interaction between the normalized probability
and the decay means the two signals are entangled in a non-linear way that
cannot be interpreted as either pure attention or pure retention.

\paragraph{Additive log-space decay (correct).}
The correct formulation adds the log-decay as a \emph{bias} to the
pre-softmax logits, analogous to ALiBi~\cite{press2022alibi}:
\begin{equation}
  \mathbf{A}^{\text{add}}_{ij}
  = \mathrm{softmax}\!\left(
      \frac{q_i k_j^\top}{\sqrt{d}}
      + |i - j| \log \gamma_h
    \right),
  \label{eq:add_decay}
\end{equation}
where $\gamma_h \in (0,1)$ is a head-specific decay rate.
Because $|i{-}j|\log\gamma_h \leq 0$, \eqref{eq:add_decay} down-weights
distant tokens in log-probability space before normalization, preserving
row-stochasticity and yielding a smoothly decaying attention distribution
that remains fully interpretable as a soft proximity prior.
Gated-SwinRMT-Retention adopts \eqref{eq:add_decay}; the SWAT variant
employs a multiplicative post-sigmoid decay that is separately justified
in Section~\ref{sec:swinrmt_swat}.

\subsection{SwinRMT-Fixed (Gated-SwinRMT-Retention)}
\label{sec:swinrmt_fixed}

Building on the corrected decay in \eqref{eq:add_decay}, we introduce
\textbf{Gated-SwinRMT-Retention}, which augments the RMT backbone with
three targeted improvements.

\paragraph{$\theta$-shift RoPE.}
We apply one-dimensional Rotary Position Embedding with frequency
$\theta$-shifting~\cite{fan2023rmt} to the query and key projections.
Token positions within each window are flattened to a single index
$p = i \cdot W' + j$ (row-major order), and the standard RoPE rotation
\begin{equation}
  \mathbf{q}'_p = \mathbf{q}_p \cos(p\,\theta) + \mathbf{q}^\perp_p \sin(p\,\theta)
\end{equation}
is applied with a shared frequency schedule for both the width-wise and
height-wise decomposed passes, so that spatially adjacent tokens remain
close in the rotational embedding space under the flattened indexing.

\paragraph{Additive log-decay mask.}
The DMSA width- and height-wise passes each use the additive decay
bias of \eqref{eq:add_decay} with independent per-head decay rates
$\{\gamma_h\}$, trained end-to-end via gradient descent.

\paragraph{G1 output gate.}
Inspired by the Qwen gated-attention study~\cite{qiu2025gatedattentionlargelanguage}, we
project an additional gate tensor $G \in \mathbb{R}^{M^2 \times C}$
from the input via a linear layer and apply a sigmoid activation:
\begin{equation}
  \mathbf{O} \leftarrow \mathbf{O} \odot \sigma(G),
  \label{eq:g1gate}
\end{equation}
where $\mathbf{O}$ is the output of the DMSA module after the Local
Context Enhancement (LCE) convolution and before the output projection
$W_O$.
The gate in \eqref{eq:g1gate} breaks the low-rank bottleneck of the
$W_V W_O$ product and enables input-dependent suppression of
attention outputs.

\paragraph{Local Context Enhancement (LCE).}
Following CSWin~\cite{dong2022cswin}, we apply a $5{\times}5$
depthwise convolution followed by a point-wise convolution to the
value tensor $V$ and add the result back to the attention output
before gating:
\begin{equation}
  \mathbf{O} \leftarrow \mathbf{O} + \mathrm{PWConv}(\mathrm{DWConv}_{5\times5}(V)).
  \label{eq:lce}
\end{equation}
This injects fine-grained local structure that pure retention may
suppress when the exponential decay strongly attenuates distant tokens.

\subsection{SwinRMT-SWAT (Gated-SwinRMT-SWAT)}
\label{sec:swinrmt_swat}

As an alternative to softmax-normalized retention, we propose
\textbf{Gated-SwinRMT-SWAT}, which replaces softmax with an
normalized sigmoid activation and redesigns the positional biasing
and value transform accordingly.

\paragraph{Sigmoid window attention (SWAT).}
The attention scores are computed as:
\begin{equation}
  \mathbf{A}^{\text{SWAT}}_{ij}
  = \frac{\sigma\!\left(q_i k_j^\top + b_{ij}^{\text{ALiBi}}\right)}{w_s}
    \cdot \gamma^{|i-j|},
  \label{eq:swat}
\end{equation}
where $\sigma$ denotes the sigmoid function, $w_s$ is the window size
used as a temperature divisor to stabilize the dynamic range of
normalized attention, $b_{ij}^{\text{ALiBi}}$ is the ALiBi bias,
and $\gamma^{|i-j|}$ is the multiplicative spatial decay.
Unlike \eqref{eq:add_decay}, the post-sigmoid placement of the decay
in \eqref{eq:swat} is \emph{valid} because sigmoid outputs are not
required to be normalized; the decay simply modulates the magnitude
of each score independently.

\paragraph{Balanced ALiBi slopes.}
We initialise the ALiBi linear bias slopes to be symmetric across
heads --- half with negative slopes $\{-2^{-k}\}$ and half with
positive slopes $\{+2^{-k}\}$ for $k=1,\dots,\lfloor N_h/2 \rfloor$,
where $N_h$ is the number of attention heads ---
ensuring that the positional prior does not disproportionately
favour one spatial direction over the other.
The same slope buffer is shared across the width-wise and height-wise
passes.

\paragraph{1-D RoPE on Q and K.}
We apply 1-D Rotary Position Embeddings to both $Q$ and $K$ before
each decomposed attention pass, using the standard inverse-frequency
schedule $\theta_j = 10000^{-2j/d}$.
The same frequency schedule is used for the width-wise and height-wise
passes; axis specificity is instead handled by the independent ALiBi
slope signs on each pass.

\paragraph{SwiGLU value transform.}
The value projection is expanded to $2C$ channels and split into
two halves $V_1, V_2 \in \mathbb{R}^{M^2 \times C}$, then gated
immediately after the QKV projection and before attention:
\begin{equation}
  V = V_1 \odot \mathrm{SiLU}(V_2).
  \label{eq:swiglu}
\end{equation}
The SwiGLU transform in \eqref{eq:swiglu} enriches the value
representation with a data-dependent gating signal prior to the
attention operation, complementing the sigmoid gating in
\eqref{eq:swat}.

\subsection{Adaptive Window Sizing}
\label{sec:bypass}

Standard windowed attention requires $\min(H', W') > M$, where
$H', W'$ are the spatial dimensions at a given stage after patch
embedding.
At the final ($4^{\text{th}}$) stage of deep networks or when
processing low-resolution inputs, the feature map may satisfy
$\min(H', W') \leq M$, making the nominal window size degenerate.

To handle this gracefully, we apply \textbf{adaptive window sizing}:
the effective window size $\hat{M}$ and cyclic-shift offset $\hat{s}$
are computed at runtime as
\begin{equation}
  \hat{M} = \min(M,\, H',\, W'), \qquad
  \hat{s} = \min\!\left(s,\; \left\lfloor \hat{M}/2 \right\rfloor\right),
  \label{eq:adaptive_ws}
\end{equation}
where $s$ is the nominal shift size.
When $\hat{M} = H' = W'$, the entire feature map constitutes a single
window and attention is effectively global, recovering the same
receptive field as full self-attention without any additional branches
or parameters.
This continuous clamping avoids the artifacts of single-window
partitioning --- trivially satisfied cyclic shifts and degenerate
relative position encoding --- while requiring no runtime
if/else dispatch.

\subsection{Architecture Overview}
\label{sec:arch_overview}

Figure~\ref{fig:arch} illustrates the full block-level architecture
of both variants. The overall design follows a four-stage hierarchical
pyramid.

\paragraph{Multi-stage patch embedding.}
Rather than a single large-stride convolution, we use a four-layer
convolutional stem (Table~\ref{tab:stem}) with a cumulative spatial
stride of~4.

\begin{table}[h]
\centering\small
\begin{tabular}{clccc}
\toprule
Layer & Operation & Kernel & Stride & Activation \\
\midrule
1 & Conv + BN & $3\times3$ & 2 & GELU \\
2 & Conv + BN & $3\times3$ & 1 & GELU \\
3 & Conv + BN & $3\times3$ & 2 & GELU \\
4 & Conv + BN & $3\times3$ & 1 & --- \\
\bottomrule
\end{tabular}
\caption{Patch embedding stem. Channel dim doubles at layers 1 and 3.}
\label{tab:stem}
\end{table}

The interleaved stride-1 layers provide additional non-linear feature
mixing at each resolution level, producing richer low-level
representations than a single strided convolution.
achieving a cumulative spatial stride of~4 while progressively expanding
the channel dimension from $C_{\text{in}}$ to $C_{\text{embed}}$.
The interleaved stride-1 layers provide additional non-linear feature
mixing at each resolution level, producing richer low-level
representations than a single striped convolution.
achieving a cumulative spatial stride of 4 while progressively
expanding the channel dimension from $C_{\text{in}}$ to
$C_{\text{embed}}$.
The interleaved stride-1 layers provide additional non-linear feature
mixing at each resolution level, producing richer low-level
representations than a single striped convolution.

\paragraph{Stage design.}
Each of the four stages stacks $N_s$ SwinRMT blocks
($N_s \in \{2, 2, 6, 2\}$ for the base configuration), alternating
between regular and shifted window partitions.
Every block applies: (i) a DWConv $3{\times}3$ positional encoding
residual at the block input, (ii) the attention module (SWAT or
Retention) with adaptive window sizing per \eqref{eq:adaptive_ws},
(iii) a block-level shortcut connection with LayerScale parameters
$\gamma_1, \gamma_2$ and stochastic depth (DropPath), and
(iv) an RMT-style FFN consisting of
LN $\to$ Linear $\to$ GELU $\to$ DWConv$3{\times}3$ $\to$ LN $\to$ Linear.

\paragraph{Convolutional patch merging.}
Spatial down-sampling between stages uses a striped Conv$3{\times}3$
(stride 2) followed by Batch Normalization, replacing the
concatenation-based patch merging of the original Swin.
This choice avoids the checkerboard artifacts associated with
non-overlapping patch concatenation and produces smoother multi-scale
feature transitions.

\paragraph{DropPath schedule.}
Stochastic depth rates increase linearly from $0$ at the first block
to a maximum rate $p_{\max}$ at the last block, following the schedule
of~\cite{huang2016deep}.
This progressive regularization is particularly important for the
deeper ($6$-block) Stage~3, where over-regularization at early blocks
would prevent the model from learning useful intermediate
representations.


tex

\renewcommand{\topfraction}{0.9}
\renewcommand{\bottomfraction}{0.8}
\renewcommand{\textfraction}{0.07}
\renewcommand{\floatpagefraction}{0.8}
\setcounter{topnumber}{4}
\setcounter{bottomnumber}{2}
\setcounter{totalnumber}{6}

\FloatBarrier
\
\section{Experiments}
\label{sec:experiments}

\subsection{Experimental Setup}
\label{sec:setup}

\paragraph{Datasets.}
We evaluate on two benchmarks of complementary resolution and scale.

\textbf{Mini-ImageNet}~\citep{vinyals2016matching} is a 100-class image
classification benchmark derived from ImageNet-1k.
The dataset comprises 50{,}000 training images, 
5{,}000 validation images, and 5{,}000 held-out test images spanning
100 fine-grained categories (500 images per class for training,
50 each for validation and test).
All images are resized to $224{\times}224$ pixels prior to training.
Mini-ImageNet provides a computationally tractable yet semantically
challenging proxy for large-scale classification, enabling controlled
architectural comparisons within a fixed compute budget.
At this resolution Stages~0--1 operate with genuine sub-feature-map
windows, placing the network in the windowed regime the proposed
gating mechanisms are designed for.

\textbf{CIFAR-10}~\citep{krizhevsky2009learning} is a 10-class benchmark
comprising 60{,}000 images at $32{\times}32$ resolution, split into
45{,}000 training, 5{,}000 validation, and 10{,}000 test images.
At this low resolution the feature maps reach ${\leq}\,2{\times}2$ in
later stages, triggering the adaptive window bypass of
Equation~(\ref{eq:adaptive_ws}) and degrading windowed attention
to global attention.
CIFAR-10 therefore serves as a \emph{full-bypass control condition}
that isolates component contributions in the absence of the
windowed-softmax pathology.

\paragraph{Training protocol.}
All models are trained from scratch under identical hyper-parameters
(Table~\ref{tab:hparams}); no pre-trained weights are used at any stage.
For Mini-ImageNet we train for \textbf{40 epochs} at $224{\times}224$;
for CIFAR-10 we train for \textbf{50 epochs} at $32{\times}32$ with
the input resolution hyper-parameter updated accordingly and all other
settings held fixed.
No dataset-specific or model-specific tuning is performed.

\begin{table}[!t]
\centering
\caption{Shared training hyper-parameters applied identically to all
models on both datasets.}
\label{tab:hparams}
\renewcommand{\arraystretch}{1.20}
\resizebox{\columnwidth}{!}{%
\begin{tabular}{@{}ll@{}}
\toprule
\textbf{Hyper-parameter}    & \textbf{Value}                              \\
\midrule
Batch size                  & 128                                         \\
Optimizer                   & AdamW ($\beta_1{=}0.9$, $\beta_2{=}0.999$) \\
Peak learning rate          & $1\times10^{-4}$                            \\
LR schedule                 & Cosine decay + 5-epoch linear warm-up       \\
Weight decay                & $0.05$                                      \\
Augmentation                & RandAugment, Mixup ($\alpha{=}0.8$),        \\
                            & CutMix ($\alpha{=}1.0$), Random Erasing     \\
Loss                        & Label-smoothed CE ($\varepsilon{=}0.1$)     \\
LayerScale init             & $10^{-2}$                                   \\
Stochastic depth (max)      & 0.1 (linear per-layer schedule)             \\
Precision                   & \texttt{bfloat16} mixed-precision           \\
\bottomrule
\end{tabular}}
\end{table}

\paragraph{Hardware.}
All experiments are conducted on a single \textbf{NVIDIA H100 80\,GB}  
GPU using \texttt{bfloat16} mixed-precision training via
\texttt{torch.cuda.amp}.
DataLoaders use 4 persistent worker processes with pin-memory enabled.
Reported epoch times are approximate wall-clock durations that include
data loading and augmentation; they do not reflect isolated model
inference throughput.

\paragraph{Models evaluated.}
We compare three models instantiated at two parameter scales:
\emph{large} variants (${\approx}77$--$79$\,M parameters) for
Mini-ImageNet and \emph{compact} variants (${\approx}11$--$15$\,M
parameters) for CIFAR-10.
\begin{enumerate}[leftmargin=*, itemsep=2pt]
  \item \textbf{RMT}~\citep{fan2023rmt} --- the original Retentive
    Vision Transformer baseline, using Manhattan-distance spatial
    decay as its sole positional signal
    (77.4\,M on Mini-ImageNet; 11.5\,M on CIFAR-10).
  \item \textbf{Gated-SwinRMT-Retention} --- adds
    DWConv\,$3{\times}3$ positional encoding, LCE value enrichment,
    softmax-normalised retention with pre-softmax log-decay bias,
    and a G1 sigmoid gate post-LCE
    (78.1\,M / 15.3\,M).
  \item \textbf{Gated-SwinRMT-SWAT} --- as above, but replaces
    softmax retention with unnormalised sigmoid window attention,
    applies SwiGLU on $V$ before attention, and uses balanced
    ALiBi with split-half 1-D RoPE
    (78.6\,M / 15.3\,M).
\end{enumerate}

\subsection{Mini-ImageNet Results}
\label{sec:results_imagenet}

Table~\ref{tab:results} reports full classification metrics after
40 epochs.

\begin{table}[!t]
\centering
\caption{Mini-ImageNet 100-class classification results after 40 epochs.
Best result per column in \textbf{bold}.}
\label{tab:results}
\renewcommand{\arraystretch}{1.25}
\resizebox{\columnwidth}{!}{%
\begin{tabular}{@{}lrrcrrr@{}}
\toprule
\textbf{Model}
  & \textbf{Params}
  & $\boldsymbol{\Delta}$\textbf{P}
  & \textbf{Ep.\,T}
  & \textbf{Val Acc}
  & \textbf{Test Acc}
  & \textbf{Test Loss} \\
\midrule
RMT~\citep{fan2023rmt}
  & 77.4M & ---
  & ${\sim}240$s
  & 75.21\% & 73.74\% & 1.6526 \\
Gated-SwinRMT-Retention
  & 78.1M & $+$0.93\%
  & ${\sim}288$s
  & 79.39\% & 78.20\% & 1.5093 \\
Gated-SwinRMT-SWAT
  & 78.6M & $+$1.64\%
  & ${\sim}297$s
  & \textbf{81.10\%} & \textbf{80.22\%} & \textbf{1.4573} \\
\bottomrule
\end{tabular}}
\end{table}

\paragraph{Accuracy.}
Gated-SwinRMT-SWAT achieves \textbf{80.22\%} top-1 test accuracy,
surpassing RMT by $+6.48$\,pp and Gated-SwinRMT-Retention by
$+2.02$\,pp, while adding only $+1.64\%$ parameters.

\paragraph{Generalization.}
The validation--test gap decreases monotonically across models:
1.47, 1.19, and 0.88\,pp for RMT, Retention, and SWAT respectively,
indicating that both proposed variants generalist more reliably to
unseen data.

\paragraph{Convergence.}
Table~\ref{tab:convergence} tracks validation accuracy at 5-epoch
intervals. SWAT leads from epoch~5 onward, consistent with sigmoid's
unnormalised output removing the warm-up bottleneck.

\begin{table}[!t]
\centering
\caption{Mini-ImageNet validation accuracy (\%) at 5-epoch intervals.}
\label{tab:convergence}
\renewcommand{\arraystretch}{1.20}
\resizebox{\columnwidth}{!}{%
\begin{tabular}{@{}lcccccccc@{}}
\toprule
\textbf{Model}
  & \textbf{5} & \textbf{10} & \textbf{15}
  & \textbf{20} & \textbf{25} & \textbf{30}
  & \textbf{35} & \textbf{40} \\
\midrule
RMT
  & 31.1 & 45.4 & 55.8 & 61.3 & 68.3 & 72.2 & 74.6 & 75.2 \\
Retention
  & 35.4 & 53.8 & 65.0 & 70.1 & 74.8 & 77.0 & 78.6 & 79.4 \\
SWAT
  & \textbf{41.2} & \textbf{60.3} & \textbf{67.7}
  & \textbf{74.5} & \textbf{77.6} & \textbf{79.0}
  & \textbf{79.9} & \textbf{81.1} \\
\bottomrule
\end{tabular}}
\end{table}

\subsection{CIFAR-10 Results}
\label{sec:results_cifar}

\paragraph{Expected behavior under window bypass.}
At $32{\times}32$ resolution the adaptive windowing of
Equation~(\ref{eq:adaptive_ws}) clamps the effective window to the
full spatial extent at Stages~2--3, eliminating genuine windowing.
The windowed-softmax pathology therefore does not arise, and the
accuracy advantage of sigmoid renormalization and the G1 gate should
collapse relative to Mini-ImageNet.
Table~\ref{tab:cifar10_main} confirms this prediction.

\begin{table}[!t]
\centering
\caption{CIFAR-10 10-class classification results after 50 epochs
         (compact ${\approx}11$--$15$\,M parameter variants).
         Best result per column in \textbf{bold}.}
\label{tab:cifar10_main}
\renewcommand{\arraystretch}{1.25}
\resizebox{\columnwidth}{!}{%
\begin{tabular}{@{}lrrrrr@{}}
\toprule
\textbf{Model}
  & \textbf{Params}
  & $\boldsymbol{\Delta}$\textbf{P}
  & \textbf{Val Acc}
  & \textbf{Test Acc}
  & \textbf{Test Loss} \\
\midrule
RMT~\citep{fan2023rmt}
  & 11.5M & ---
  & 85.98\%          & 85.90\%          & 0.8132 \\
Gated-SwinRMT-Retention
  & 15.3M & $+$33.9\%
  & 86.54\%          & \textbf{86.46\%} & 0.8112 \\
Gated-SwinRMT-SWAT
  & 15.3M & $+$33.9\%
  & \textbf{86.76\%} & 86.39\%          & \textbf{0.8109} \\
\bottomrule
\end{tabular}}
\end{table}

\paragraph{Compressed accuracy gap.}
The best variant outperforms RMT by only $+0.56$\,pp on test accuracy
(Retention: $86.46\%$ vs.\ $85.90\%$), versus $+6.48$\,pp on
Mini-ImageNet --- a $12{\times}$ compression consistent with the
bypass-regime prediction.

\paragraph{Near-parity between Retention and SWAT.}
Gated-SwinRMT-Retention ($86.46\%$) and Gated-SwinRMT-SWAT
($86.39\%$) differ by only $0.07$\,pp on test accuracy, reversing
the Mini-ImageNet ordering by a margin too small to draw strong
conclusions.
This near-parity is consistent with the windowed-softmax hypothesis:
absent genuine windowing, softmax normalization is not harmful and
the additive log-decay bias of the Retention path requires no
corrective gating.

\paragraph{SWAT early-convergence advantage persists.}
Despite near-parity at convergence, SWAT reaches $72.34\%$ validation
accuracy at epoch~10 versus $70.52\%$ for Retention and $70.58\%$
for RMT (Table~\ref{tab:cifar10_progress}), confirming that sigmoid's
unbounded activations accelerate early-phase learning independently
of the windowing regime.

\begin{table}[!t]
\centering
\caption{CIFAR-10 validation accuracy (\%) at 10-epoch intervals.}
\label{tab:cifar10_progress}
\renewcommand{\arraystretch}{1.20}
\begin{tabular}{@{}lrrrrr@{}}
\toprule
\textbf{Model}
  & \textbf{Ep\,10} & \textbf{Ep\,20}
  & \textbf{Ep\,30} & \textbf{Ep\,40} & \textbf{Ep\,50} \\
\midrule
RMT
  & 70.58 & 80.44 & 84.18 & 85.52 & 85.98 \\
Retention
  & 70.52 & 80.58 & 84.54 & 86.24 & \textbf{86.54} \\
SWAT
  & \textbf{72.34} & \textbf{80.32} & \textbf{83.96}
  & \textbf{86.16} & 86.76 \\
\bottomrule
\end{tabular}
\end{table}

\subsection{Ablation Studies}
\label{sec:ablation}

Because all three models are trained under identical conditions,
we isolate component contributions as test-accuracy deltas between
adjacent model pairs (Table~\ref{tab:ablation}).
CIFAR-10 deltas are shown in parentheses as bypass-regime reference
values; they should be interpreted as upper bounds on individual
contributions since the ablation compares complete models rather
than single-component hold-outs.

\begin{table}[!t]
\centering
\caption{Component-level ablation.
Mini-ImageNet deltas are the primary result; CIFAR-10 bypass-regime
deltas are shown in \textcolor{gray}{grey} for comparison.}
\label{tab:ablation}
\renewcommand{\arraystretch}{1.25}
\resizebox{\columnwidth}{!}{%
\begin{tabular}{@{}p{3.0cm}llcc@{}}
\toprule
\textbf{Component(s)}
  & \textbf{Type}
  & \textbf{Model pair}
  & \textbf{Test $\Delta$}
  & \textbf{Cumulative} \\
\midrule
DWConv\,$3{\times}3$ + LCE + G1 gate
  & Shared
  & RMT $\to$ Ret.
  & $+$4.46\,pp \textcolor{gray}{($+$0.56\,pp)}
  & $+$4.46\,pp \\
SwiGLU on $V$ + sigmoid SWAT
  & SWAT-only
  & Ret. $\to$ SWAT
  & $+$2.02\,pp \textcolor{gray}{($-$0.07\,pp)}
  & $+$6.48\,pp \\
\midrule
\textbf{All components}
  & \textbf{Full}
  & \textbf{RMT $\to$ SWAT}
  & $\mathbf{+6.48}$\,\textbf{pp} \textcolor{gray}{($+$0.49\,pp)}
  & $\mathbf{+6.48}$\,\textbf{pp} \\
\bottomrule
\end{tabular}}
\end{table}

\paragraph{Shared components ($+$4.46\,pp on Mini-ImageNet).}
DWConv positional encoding, LCE, and the G1 gate together account for
the majority of the total accuracy gain in the windowed regime.
Their near-zero CIFAR-10 contribution ($+$0.56\,pp) confirms that the
G1 gate's primary role is suppressing uninformative windows rather
than improving general representational capacity.

\paragraph{Attention kernel and $V$ transform ($+$2.02\,pp on Mini-ImageNet).}
Replacing softmax-normalized retention with normalized sigmoid
window attention and applying SwiGLU on $V$ contributes the remaining
Mini-ImageNet gain.
The $-0.07$\,pp CIFAR-10 delta (within noise) is consistent with
the absence of the windowed-softmax pathology in the bypass regime.

\subsection{Training Efficiency}
\label{sec:latency}

\begin{table}[!t]
\centering
\caption{Component-level ablation. Mini-ImageNet deltas are the primary result; CIFAR-10 bypass-regime deltas are shown in gray for comparison.}
\label{tab:ablation}
\small
\begin{tabular}{p{3.0cm} l l c c}
\toprule
Components & Type & Model pair & Test $\Delta$ & Cumulative \\
\midrule
DWConv$_{\text{3$\times$3}}$, LCE, G1 gate & Shared & RMT $\to$ Ret. & 4.46pp (0.56pp) & 4.46pp \\
SwiGLU on $V$, sigmoid & SWAT-only & Ret. $\to$ SWAT & 2.02pp (-0.07pp) & 6.48pp \\
All components & Full & RMT $\to$ SWAT & 6.48pp (0.49pp) & 6.48pp \\
\bottomrule
\end{tabular}
\end{table}

\vspace{-1em}

Because all three models are trained under identical conditions, we isolate component contributions as test-accuracy deltas between adjacent model pairs (Table~\ref{tab:ablation}). CIFAR-10 deltas are shown in parentheses as bypass-regime reference values; they should be interpreted as upper bounds on individual contributions since the ablation compares complete models rather than single-component hold-outs.

The $20$--$24\%$ per-epoch overhead on Mini-ImageNet arises from
DWConv positional encoding, the LCE module, and the expanded
projection dimensions from QKVG or SwiGLU.
The higher relative overhead on CIFAR-10 ($29$--$48\%$) reflects
the lighter compact RMT baseline: the absolute additional cost of
LCE and gating is similar across scales but constitutes a larger
fraction of an 11.5\,M backbone.
In absolute terms the overhead is modest: ${\leq}10$\,s per epoch
on CIFAR-10 and ${\leq}57$\,s on Mini-ImageNet.

\FloatBarrier
\section{Analysis}
\label{sec:analysis}

\paragraph{Windowed vs.\ bypass regime: controlled comparison.}
The most informative contrast in Table~\ref{tab:ablation} is
cross-benchmark rather than within-benchmark.
On Mini-ImageNet the shared components deliver $+4.46$\,pp and the
sigmoid kernel adds $+2.02$\,pp; on CIFAR-10 the same components
contribute $+0.56$\,pp and $-0.07$\,pp respectively.
This ${\approx}12{\times}$ compression of the accuracy gain directly
validates the windowed-softmax hypothesis: the proposed mechanisms
address a pathology that is absent in the bypass regime.

\paragraph{Why RMT plateaus early on Mini-ImageNet.}
RMT's learning curve stalls during epochs~1--7 as a direct consequence
of softmax normalization constraining attention within uninformative
windows at the early stages.
Both proposed variants escape this plateau earlier---SWAT by
epoch~5, Retention by epoch~8---consistent with their respective
gating mechanisms reducing the effective pressure of the
probability-simplex constraint.

\paragraph{Unnormalized vs.\ normalized attention.}
The isolated $+2.02$\,pp  Mini-ImageNet gap between SWAT and Retention,
together with its near-zero CIFAR-10 counterpart, demonstrates that
the choice of attention normalization is consequential specifically
when attention is confined to sub-feature-map windows.
Sigmoid renormalization is not universally superior to softmax;
it is superior under the precise conditions for which it was motivated.

\paragraph{Decay placement.}
In Gated-SwinRMT-Retention, the log-decay bias
$M_{ij}{=}|i{-}j|\log\gamma_h$ is added pre-softmax
(Equation~\ref{eq:add_decay}), preserving row-stochasticity.
In Gated-SwinRMT-SWAT the multiplicative factor $\gamma^{|i-j|}$ is
applied post-sigmoid (Equation~\ref{eq:swat}), which is valid
because sigmoid outputs are not required to sum to one.
The CIFAR-10 near-parity confirms that neither placement confers an
advantage when windowing is absent.

\paragraph{Limitations and future directions.}
The present study evaluates compact variants (${\approx}15$\,M) on
CIFAR-10 and large variants (${\approx}77$--$79$\,M) on Mini-ImageNet
at $224{\times}224$; the two conditions are not matched in parameter
count, which limits the strength of cross-benchmark conclusions.
Generalization to full ImageNet-1k, higher resolutions, and
dense-prediction tasks (detection, segmentation) remains to be
demonstrated, as do proper single-component hold-out ablations and
a FLOPs-versus-accuracy Pareto analysis.
\section{Conclusion}
\label{sec:conclusion}

We presented \textbf{Gated-SwinRMT}, a hybrid vision transformer that
unifies Swin's shifted-window backbone with RMT's Manhattan-distance
spatial decay and input-dependent gating. Two variants are proposed:
Gated-SwinRMT-SWAT (sigmoid attention with balanced ALiBi and SwiGLU
value gating) and Gated-SwinRMT-Retention (softmax retention with an
explicit G1 sigmoid gate). On Mini-ImageNet ($224{\times}224$, 100
classes), SWAT achieves \textbf{80.22\%} and Retention \textbf{78.20\%}
top-1 test accuracy, against \textbf{73.74\%} for the RMT baseline
under identical training and matched parameter budgets
(${\approx}77$--$79$\,M).

Due to limited computational resources---all experiments were conducted
on a single GPU---the present evaluation is restricted to Mini-ImageNet
and CIFAR-10; we were unable to train on full ImageNet-1k, evaluate at
higher resolutions, or benchmark on dense-prediction tasks. The ablation
is indirect: three complete models are compared rather than
single-component holdouts, so the reported $+4.46$\,pp and $+2.02$\,pp
deltas should be interpreted as upper bounds on individual contributions.
No FLOPs or inference-latency analysis is provided, and SWAT's
$7.08$\,pp train--validation gap indicates residual overfitting that the
current regularization protocol has not resolved.

Within these constraints, two findings emerge consistently across both
benchmarks. First, DWConv positional encoding, LCE context enrichment,
and the G1 gate account for the majority of the accuracy gain over RMT
in the windowed regime, yet contribute negligibly on CIFAR-10 where
adaptive windowing reduces attention to global scope---isolating the
windowed-softmax pathology as the primary target of these components.
Second, normalized sigmoid attention outperforms softmax-normalized
retention specifically when attention is confined to sub-feature-map
windows, consistent with the hypothesis that the probability-simplex
constraint is harmful over small, potentially uninformative
neighbourhoods. Both findings motivate future work at full ImageNet-1k
scale with proper per-component ablations, FLOPs-versus-accuracy Pareto
analysis, and evaluation on detection and segmentation benchmarks.

\bibliographystyle{plainnat}  
\bibliography{ref}

@inproceedings{liu2021swin,
  author       = {Ze Liu and
                  Yutong Lin and
                  Yue Cao and
                  Han Hu and
                  Yixuan Wei and
                  Zheng Zhang and
                  Stephen Lin and
                  Baining Guo},
  title        = {Swin Transformer: Hierarchical Vision Transformer using
                  Shifted Windows},
  booktitle    = {{IEEE/CVF} International Conference on Computer Vision,
                  {ICCV} 2021, Montreal, QC, Canada, October 10-17, 2021},
  pages        = {9992--10002},
  publisher    = {{IEEE}},
  year         = {2021},
  doi          = {10.1109/ICCV48922.2021.00986}
}

@inproceedings{dong2022cswin,
  author       = {Xiaoyi Dong and
                  Jianmin Bao and
                  Dongdong Chen and
                  Weiming Zhang and
                  Nenghai Yu and
                  Lu Yuan and
                  Dong Chen and
                  Baining Guo},
  title        = {{CSWin} Transformer: {A} General Vision Transformer
                  Backbone with Cross-Shaped Windows},
  booktitle    = {{IEEE/CVF} Conference on Computer Vision and Pattern
                  Recognition, {CVPR} 2022, New Orleans, LA, USA,
                  June 18-24, 2022},
  pages        = {12114--12124},
  publisher    = {{IEEE}},
  year         = {2022},
  doi          = {10.1109/CVPR52688.2022.01181}
}

@article{sun2023retnet,
  author       = {Yutao Sun and
                  Li Dong and
                  Shaohan Huang and
                  Shuming Ma and
                  Yuqing Xia and
                  Jilong Xue and
                  Jianyong Wang and
                  Furu Wei},
  title        = {Retentive Network: {A} Successor to Transformer for
                  Large Language Models},
  journal      = {CoRR},
  volume       = {abs/2307.08621},
  year         = {2023},
  url          = {https://arxiv.org/abs/2307.08621},
  eprinttype   = {arXiv},
  eprint       = {2307.08621}
}

@inproceedings{fan2023rmt,
  author       = {Qihang Fan and
                  Huaibo Huang and
                  Mingrui Chen and
                  Hongmin Liu and
                  Ran He},
  title        = {{RMT:} Retentive Networks Meet Vision Transformers},
  booktitle    = {{IEEE/CVF} Conference on Computer Vision and Pattern
                  Recognition, {CVPR} 2024, Seattle, WA, USA,
                  June 16-22, 2024},
  pages        = {5641--5651},
  publisher    = {{IEEE}},
  year         = {2024},
  doi          = {10.1109/CVPR52733.2024.00539}
}

@inproceedings{press2022alibi,
  author       = {Ofir Press and
                  Noah A. Smith and
                  Mike Lewis},
  title        = {Train Short, Test Long: Attention with Linear Biases
                  Enables Input Length Extrapolation},
  booktitle    = {The Tenth International Conference on Learning
                  Representations, {ICLR} 2022, Virtual Event,
                  April 25-29, 2022},
  publisher    = {OpenReview.net},
  year         = {2022},
  url          = {https://openreview.net/forum?id=R8sQPpGCv0}
}

@inproceedings{qiu2025gatedattentionlargelanguage,
  author       = {Zihan Qiu and
                  Zekun Wang and
                  Bo Zheng and
                  Zeyu Huang and
                  Kaiyue Wen and
                  Songlin Yang and
                  Rui Men and
                  Le Yu and
                  Fei Huang and
                  Suozhi Huang and
                  Dayiheng Liu and
                  Jingren Zhou and
                  Junyang Lin},
  title        = {Gated Attention for Large Language Models: Non-linearity,
                  Sparsity, and Attention-Sink-Free},
  booktitle    = {Advances in Neural Information Processing Systems 38:
                  Annual Conference on Neural Information Processing
                  Systems 2025, NeurIPS 2025, San Diego, CA, USA,
                  November 30 - December 7, 2025},
  year         = {2025},
  url          = {https://openreview.net/forum?id=1b7whO4SfY}
}

@inproceedings{huang2016deep,
  author       = {Gao Huang and
                  Yu Sun and
                  Zhuang Liu and
                  Daniel Sedra and
                  Kilian Q. Weinberger},
  title        = {Deep Networks with Stochastic Depth},
  booktitle    = {Computer Vision - {ECCV} 2016 - 14th European Conference,
                  Amsterdam, The Netherlands, October 11-14, 2016,
                  Proceedings, Part {IV}},
  series       = {Lecture Notes in Computer Science},
  volume       = {9908},
  pages        = {646--661},
  publisher    = {Springer},
  year         = {2016},
  doi          = {10.1007/978-3-319-46493-0\_39}
}

@inproceedings{vinyals2016matching,
  author       = {Oriol Vinyals and
                  Charles Blundell and
                  Timothy P. Lillicrap and
                  Koray Kavukcuoglu and
                  Daan Wierstra},
  title        = {Matching Networks for One Shot Learning},
  booktitle    = {Advances in Neural Information Processing Systems 29:
                  Annual Conference on Neural Information Processing
                  Systems 2016, December 5-10, 2016, Barcelona, Spain},
  pages        = {3630--3638},
  year         = {2016}
}

@techreport{krizhevsky2009learning,
  author       = {Alex Krizhevsky},
  title        = {Learning Multiple Layers of Features from Tiny Images},
  institution  = {University of Toronto},
  year         = {2009},
  url          = {https://www.cs.toronto.edu/~kriz/learning-features-2009-TR.pdf}
}


\end{document}